\documentclass[letterpaper, 10 pt, conference]{ieeeconf}
\IEEEoverridecommandlockouts
\let\labelindent\relax
\usepackage{enumitem}
\usepackage{balance}
\usepackage[font={footnotesize}]{caption}
\usepackage{caption}
\usepackage{subcaption}
\usepackage{array}
\usepackage{textcomp}
\usepackage{mathtools, nccmath}
\usepackage{hyperref}
\usepackage{graphicx}
\usepackage{amsfonts}
\usepackage{amsmath}
\usepackage{amssymb}
\usepackage{algorithm}
\usepackage{algorithmicx}
\usepackage{tikz}
\usepackage{arydshln}
\usepackage{multirow}
\usepackage{bm}
\usepackage{epstopdf}
\usepackage{siunitx}
\usepackage{xcolor}
\usepackage{tabularx}
\usepackage{cite}
\usepackage{listings}
\usepackage{bbold}
\usepackage{booktabs}      
\usepackage{adjustbox}

\makeatletter
\let\NAT@parse\undefined
\makeatother
\usepackage{hyperref}  
\title{\LARGE \textbf{Learning Generalizable Language-Conditioned Cloth Manipulation from Long Demonstrations}}

\author{
Hanyi Zhao$^{*}$, Jinxuan Zhu$^{*}$, Zihao Yan$^{*}$, Yichen Li, Yuhong Deng, and Xueqian Wang
\thanks{$^*$ indicates equal contribution.}
\thanks{Correspondence to: {\tt\small wang.xq@sz.tsinghua.edu.cn}}
\thanks{Hanyi Zhao, Jinxuan Zhu, Zihao Yan, Yichen Li, and Xueqian Wang are with the Center for Artificial Intelligence and Robotics, Tsinghua  Shenzhen  International Graduate School, Shenzhen, China.}
\thanks{
Yuhong Deng is with School of Computing, National University of Singapore, Singapore. }
}

\begin{document}
\maketitle

\begingroup
\begin{abstract}
Multi-step cloth manipulation is a challenging problem for robots due to the high-dimensional state spaces and the dynamics of cloth. Despite recent significant advances in end-to-end imitation learning for multi-step cloth manipulation skills, these methods fail to generalize to unseen tasks. Our insight in tackling the challenge of generalizable multi-step cloth manipulation is decomposition. We propose a novel pipeline that autonomously learns basic skills from long demonstrations and composes learned basic skills to generalize to unseen tasks. Specifically, our method first discovers and learns basic skills from the existing long demonstration benchmark with the commonsense knowledge of a large language model (LLM). Then, leveraging a high-level LLM-based task planner, these basic skills can be composed to complete unseen tasks.
Experimental results demonstrate that our method outperforms baseline methods in learning multi-step cloth manipulation skills for both seen and unseen tasks. Project website: 
\href{https://sites.google.com/view/gen-cloth}{https://sites.google.com/view/gen-cloth}

\end{abstract}
\endgroup

\section{Introduction}
\label{sec:introduction}

Cloth manipulation skills like folding the T-shirt are crucial for intelligent household robots, and most cloth manipulation tasks require multiple sequential actions to achieve a desired state. 
However, multi-step cloth manipulation has posed new challenges for robot manipulation. Due to the deformable nature of cloth, it has high-dimensional state spaces and complex non-linear dynamics~\cite{zhu2022challenges}. As a result, the state of cloth is hard to predict in multi-step manipulation. A slightly different interaction may lead to significantly different cloth behaviors.
Furthermore, in multi-step cloth manipulation, a wrong action can cause the cloth to crumple, which is irreversible~\cite{gctn}. Consequently, successfully completing the task requires a particular sequence of actions. \par 

For multi-step cloth manipulation tasks, early works used heuristic policies with task-specific grippers based on physical models~\cite{doumanoglou2016folding, bersch2011bimanual}. These methods lack generalization and struggle with distractions during execution. More recently, learning-based methods have made significant advances in multi-step cloth manipulation by learning robot actions directly from expert demonstrations~\cite{seita2020deep, salhotra2022learning}. However, most of these methods try to learn task-specific multi-step manipulation policies and often struggle to generalize to unseen tasks. In this paper, we develop a learning method for generalizable cloth manipulation, which can perform well on unseen multi-step manipulation tasks.

\begin{figure}
    \centering
    \includegraphics[width=\linewidth]{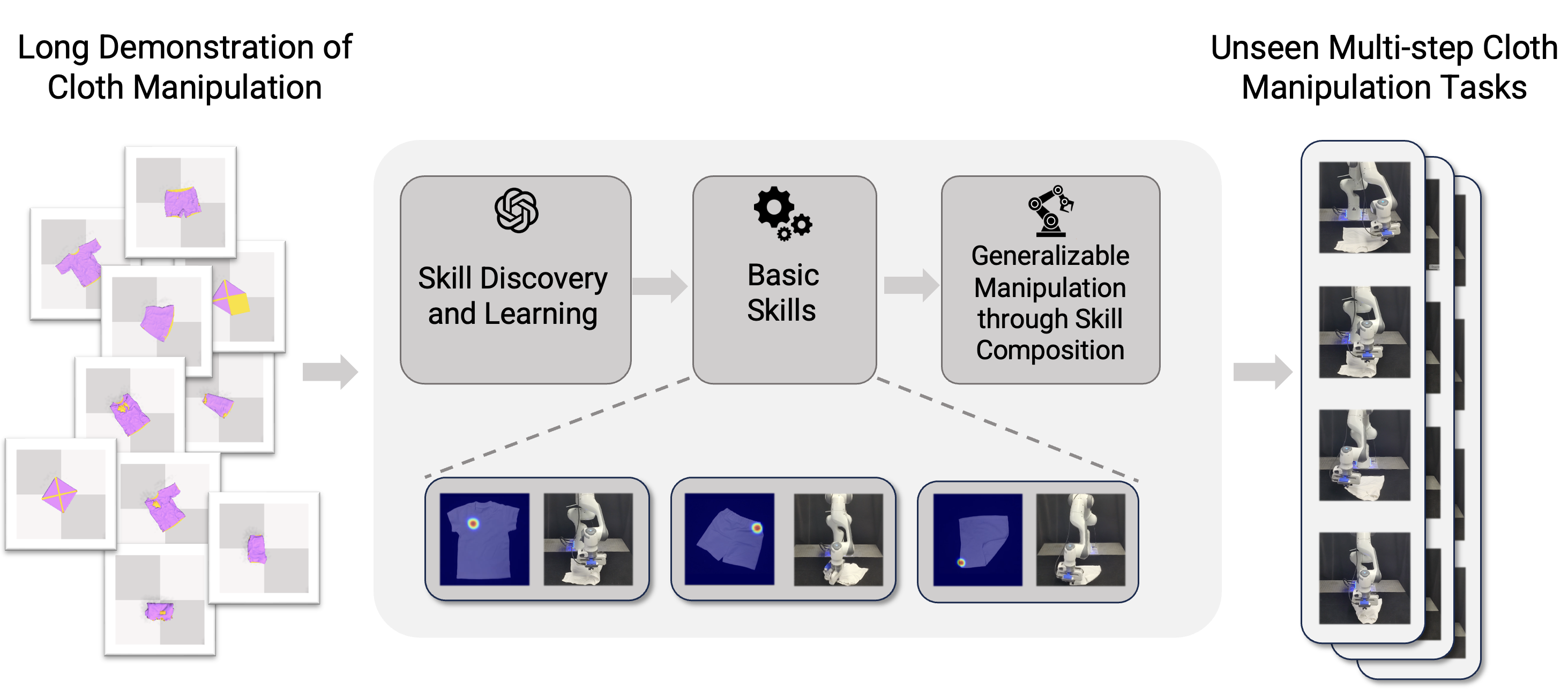}
    \caption{\textbf{Generalizable multi-step cloth manipulation}. The proposed method can learn generalizable basic skills from long demonstrations and generalize to unseen multi-step cloth manipulation tasks.}
    \label{fig:task}
    \vspace{-0.6cm}
\end{figure}

Considering the challenges and complexity of multi-step cloth manipulation, our key idea is decomposing the multi-step cloth manipulation task into generalizable basic skills and subsequently address multi-step, unseen tasks by composing these learned basic skills (Fig.~\ref{fig:task}). Unlike rigid objects with well-defined basic actions such as grasping, moving, and placing, cloth manipulation lacks basic skills. A significant gap
remains between task planning and action execution. So we first utilize an LLM to autonomously discover basic skills from existing multi-step cloth manipulation benchmarks. In this stage, an LLM is prompted to discover generalizable basic skills from long demonstrations (e.g., \textit{``pick up the left sleeve''} from the task \textit{``fold the t-shirt for storage''}). Based on the skill discovery results, the existing multi-step cloth manipulation benchmark can be broken down into a dataset for language-conditioned basic skill learning. Inspired by previous work~\cite{languagedeformable}, we then leverage a transformer-based model architecture to learn language-conditioned policies from the established basic skill dataset. After obtaining these language-conditioned basic skills, we leverage an LLM-based task planner to compose learned basic skills for unseen multi-step cloth manipulation tasks. Like previous work in rigid object manipulation~\cite{liang2023code, yu2024octopi}, the LLM-based task planner can provide commonsense
knowledge for reasoning, significantly enhancing generalization capabilities.

To evaluate the proposed method, we first conducted simulation experiments in SoftGym benchmark~\cite{lin2021softgym}, including 10 multi-step cloth manipulation tasks across 5 common cloth categories. The proposed method outperforms baseline methods in both seen and unseen tasks. Real-world experiments show that our method trained in simulation can be easily transferred to the real world. These results demonstrate that the proposed method is effective for generalizable multi-step cloth manipulation.

In summary, our main contributions are as follows:
\begin{itemize}
    \item We propose a novel pipeline that learns basic skills from long demonstrations and composes learned basic skills to generalize to unseen multi-step cloth manipulation tasks.
    \item We propose an autonomous basic skill discovery method for cloth manipulation, where an LLM is used to provide
    commonsense knowledge.
    \item We conduct extensive experiments to validate the effectiveness and generalization of our proposed pipeline in both simulation and real-world environments.
\end{itemize}

\section{Related Work}
\label{sec:related}

\subsection{Deformable Object Manipulation}

There are two primary approaches for deformable object manipulation: model-based and learning-based methods. Specifically, model-based techniques rely on a dynamic model to predict changes in the configuration of a deformable object and select the appropriate actions~\cite{hu2018three,cusumano2011bringing}. Additionally, some studies have employed the concept of diminishing rigidity~\cite{withoutmodeling} or particle-based representations~\cite{chen2024differentiable} to address the challenge of accurate system modeling. However, the computational complexity and the unavoidable accumulation of errors with increasing step size make them less effective on multi-step cloth manipulation tasks.

In contrast, learning-based methods aim to learn robot actions directly from expert demonstrations without a dynamic model. However, these approaches are typically designed for specific tasks and lack generalizability~\cite{foldsformer,yu2022global,xu2022dextairity}. Recent advancements have introduced language conditioning~\cite{languagedeformable} and semantic keypoints~\cite{deng2024clasp} to facilitate generalizable manipulation. These approaches can adapt to diverse language instructions and relatively short-horizon tasks, but are still limited to task-specific action sequences. 
In this paper, our method focuses on learning decomposed single-step skills, which are not only easier to acquire but also reusable, enabling generalization to unseen multi-step tasks.

\subsection{Foundation Models for Robot Manipulation}

Foundation models have recently been extensively studied for robot manipulation. Some works leverage internet-scale data, combined with cross-embodiment robotic data, to train end-to-end vision-language-action models (VLAs)~\cite{black2024pi_0,kim2024openvla}, demonstrating significant potential in terms of generalizability and semantic understanding. However, these models still struggle with multi-step, complex tasks without fine-tuning and are not yet mature enough for direct deployment.
Other works employ pre-trained large language models (LLMs) or large vision-language models (VLMs) for planning and perception in robotic manipulation, demonstrating substantial enhancement and adaptability. For instance, various foundation models have been extensively utilized for generating executable code~\cite{liang2023code, xiao2024robi}, formulating high-level plans~\cite{huang2022language, chen2023llm, manual2skill}, and supplying commonsense knowledge and perceptual information to enhance classical planners~\cite{luo2024gson, di2024dinobot, tang2025functo}.
However, previous methods are limited to rigid objects and focus primarily on recomposing or planning primitive actions. Instead of solely recomposing primitive actions, we propose a framework that autonomously discovers basic skills from existing benchmarks.

\subsection{Skill Discovery}

Skill discovery from demonstrations has emerged as a vital research direction in robotic manipulation, aiming to discover reusable and modular sub-skills from existing datasets. Previous works often use unsupervised skill discovery methods~\cite{skilldiscoverywithMI3,skilldiscoverywithWD1}. However, these approaches face limitations in their segmented understanding of skills, which makes them less effective for reusability and modularity.
Some works~\cite{zhu2022bottom, chu2019real, pirk2020modeling} focus on discovering reusable skills from unsegmented demonstrations for adaptive manipulation. Recent advancements~\cite{rho2024language, liulearning} leverage the semantic knowledge of LLMs for skill discovery, which enhances the semantic diversity of the resulting skills and provides a simpler way to utilize learned skills through natural language. Inspired by these advancements, and given the lack of single-step basic skill datasets for cloth manipulation in prior works, our method leverages an LLM to discover skills from long demonstrations of cloth manipulation and compose the skills for unseen tasks.

 \begin{figure*}[htbp]
    \centering
    \includegraphics[width=\linewidth]{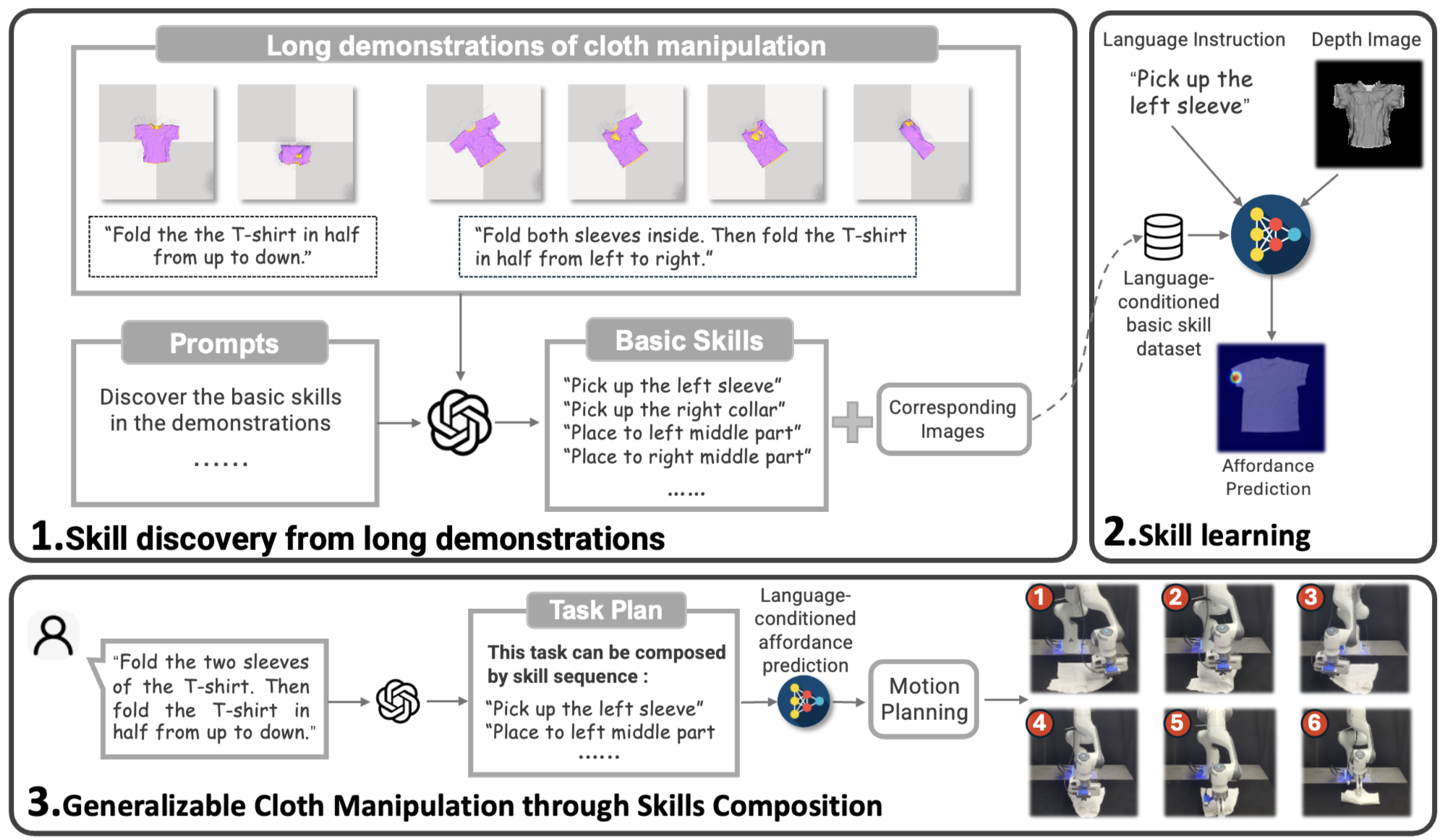}
    \caption{\textbf{Method overview.} The proposed framework consists of three stages. First, we perform skill discovery from long demonstrations and establish a language-conditioned basic skill dataset. The established dataset will then be used to train the basic skills. Finally, an LLM-based task planner will be used to compose the basic skills learned for unseen multi-step manipulation tasks. }
    \label{fig:method_overview}
    \vspace{-0.6cm}
\end{figure*}

\section{Problem Formulation}
We aim to learn basic language-conditioned cloth manipulation skills from existing multi-step manipulation datasets. Specifically, given a dataset of language-conditioned cloth manipulation demonstrations from prior work \cite{languagedeformable}, denoted as $D = \{D_1, D_2, \dots, D_N\}$, where each $D_i$ represents a manipulation task comprising a sequence of discrete depth images $I_i = \{I_{i1}, I_{i2}, \dots, I_{im}\}$, a corresponding natural language instruction $L_i$ (e.g., \textit{``Fold both sleeves inside, then fold the T-shirt in half from left to right"}), and action steps $A_i = \{a_{i1}, a_{i2}, \dots, a_{i(2m-3)}, a_{i(2m-2)}\}$. Each image is associated with a pick-and-place action pair, except for the last image; thus, the total number of actions is $2m - 2$. Our objective is to learn a generalizable policy $\pi_{\theta}$ capable of generating actions based on current observations and an unseen language instruction $L_{\text{new}}$, which specifies a cloth manipulation task not included in the training demonstrations.

The main idea of our method is to perform skill discovery to learn basic language-conditioned skills from the long demonstrations using an LLM and try to compose the learned skills to solve unseen tasks.
Given the long demonstrations, we discover basic skills through the decomposition of the demonstration and instruction, using our automatic decomposition strategy $\Gamma$: 

\begin{equation}
\Gamma(D_i) \rightarrow \left\{
\begin{aligned}
&(I_{i1}, l_{i1}, a_{i1}), \\
&(I_{i1}, l_{i2}, a_{i2}), \\
&\ldots, \\
&(I_{i(m-1)}, l_{i(2m-3)}, a_{i(2m-3)}), \\
&(I_{i(m-1)}, l_{i(2m-2)}, a_{i(2m-2)})
\end{aligned}
\right\}
\end{equation}

For each $I_{ik} (k = 1, 2, \dots, m-1)$, there exist two corresponding language instructions $l_{ij}$ and action steps $a_{ij}(j = 1, 2, \dots, 2m-3, 2m-2)$, specifically $l_{i(2k-1)}$, $l_{i(2k)}$ and $a_{i(2k-1)}$, $a_{i(2k)} $. $l_{ij}$ represents a single-step language instruction indicating a ``pick'' or ``place'' action such as \textit{``pick up the right collar''} and $a_{ij}$ is the associated action.

Given such decomposed demonstrations, we aim to train a model $A$ that infers target points from language instructions and observations
$A(o_t,l_t)\to p_t$.
Given an unseen task $L_{\text{new}}$, our policy $\pi_{\theta}$ first uses a high-level planner $P$ to generate single-step language instructions that could finish $L_{\text{new}}$:
$P(L_{\text{new}}) \to \{ l_1, l_2, \dots, l_T \}$
and then apply the learned model $A$ to map all single-step language instructions $l_t$ and observations $o_t$ to the target point $p_t$.

\section{Method}
\label{sec:method}

In the following subsections, we will present how we discover and learn basic skills from long demonstrations in Subsection~\ref{subsec:task_planning}. Following that, we will introduce how we generalize to unseen multi-step tasks in Subsection~\ref{subsec:action_generation}.

\subsection{Overall Framework}

The proposed framework consists of three stages, as illustrated in Fig.~\ref{fig:method_overview}. First, we perform skill discovery from long demonstrations by inputting $N$ language instructions $L_i$ from long cloth manipulation demonstrations $D$ along with prompts to the LLM, generating a basic skill set composed of ``pick" or ``place" instructions $l_{ij}$. Together with the corresponding depth images $I_{ik}$, these form a language-conditioned basic skill dataset. Second, skills are learned by model $A$ using the aforementioned dataset, enabling it to produce affordance predictions based on depth image observations and language instructions. Third, for generalization to unseen tasks, we first decompose the given language instruction $L_{new}$ into a sequence of basic skills with LLM planner $P$. Model $A$ then performs language-conditioned affordance predictions to determine sequential pick and place points. Finally, these actions are executed step by step through motion planning.

\subsection{Skill Discovery and Learning from Long Demonstration}

\textbf{Decompose Long Demonstration through Large Language Model:} Every multi-step task demonstration $D_i$ consists of a series of depth images $I_i$, action steps $A_i$ and corresponding long language instructions $L_i$. Since the observation of deformable objects changes once after each pick-and-place operation, each depth image $I_{ik}$ can be directly associated with ``pick" and ``place" action steps $a_{ij}$. The basic skills we want to discover are the ``pick" and ``place" language instructions $l_{ij}$ corresponding to the action $a_{ij}$. So our goal is to decompose these language instructions $l_{ij}$ from the implicit long language instruction $L_i$, which is naturally suitable for LLMs with its strong language reasoning ability. \par

To achieve effective task decomposition, we adopt few-shot prompting~\cite{achiam2023gpt} and chain-of-thought (CoT) reasoning~\cite{CoT} to generate more reliable responses. To ensure alignment between the generated responses and the ``pick" and ``place" actions, we structure the responses using predefined instruction templates: \textit{``Pick up the \{which\} of the \{cloth type\}''} and \textit{``Fold it to the \{which\}''}, where \textit{``cloth type''} represents a predefined list of fabric names in our tasks, and \textit{``which''} denotes commonly used parts of each fabric. 

\begin{figure}[htbp]
    \centering
    \includegraphics[width=\linewidth]{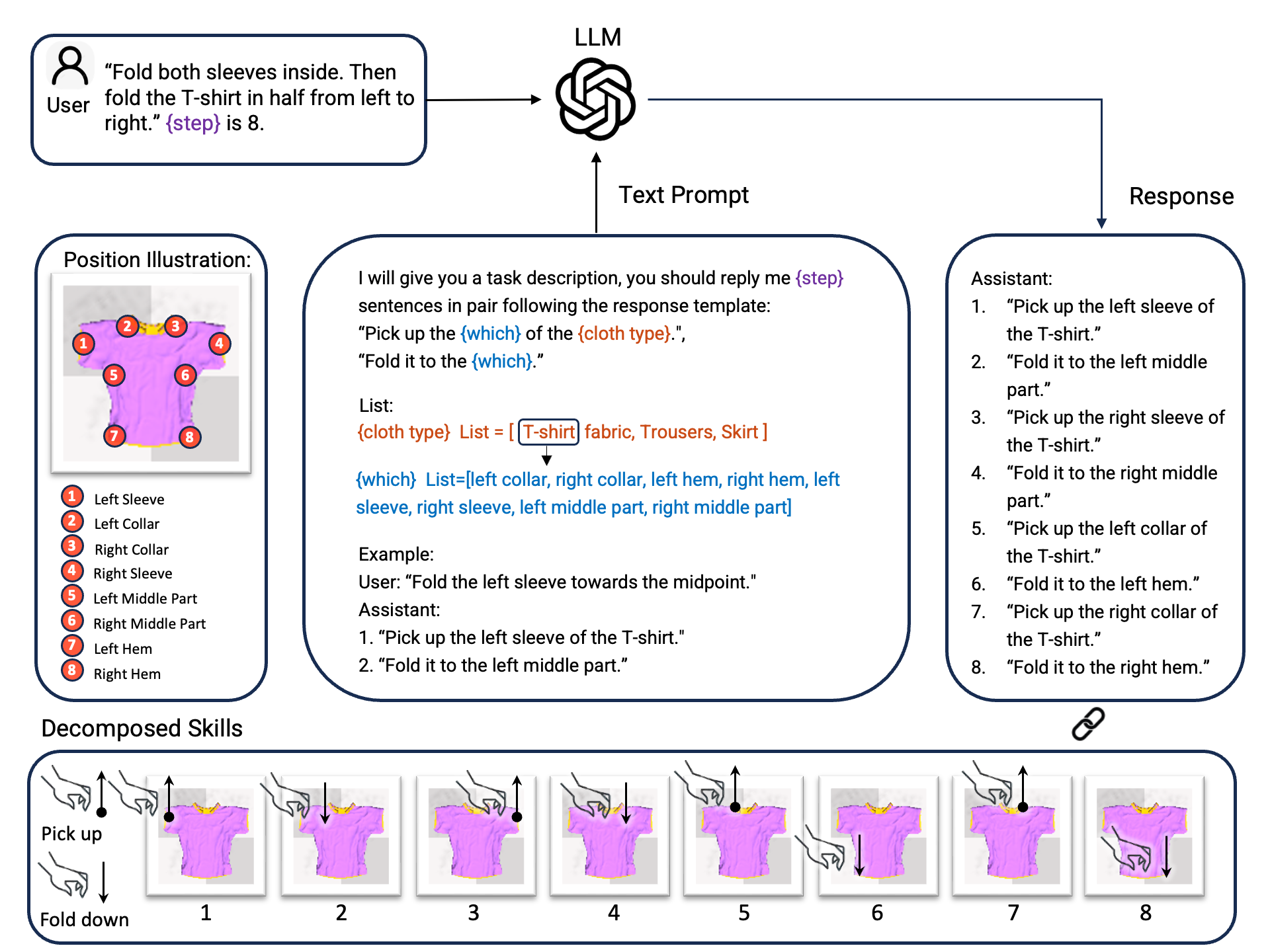}
    \caption{\textbf{Autonomous basic skill discover}. We prompt an LLM to discover basic skills from long demonstrations. }
    \label{fig:prompt}
    \vspace{-0.4cm}
\end{figure}  

As shown in Fig.~\ref{fig:prompt}, the language instruction, along with its corresponding action steps and textual prompt, is provided as input to the LLM, which then automatically decomposes the task into a sequence of ``pick" and ``place" instructions. Each instruction is associated with the observation image, which forms a language-conditioned basic skill dataset and facilitates subsequent skill learning.

Based on the diverse range of tasks in long demonstrations, we can acquire various fabric manipulation skills that encompass the majority of ``pick" and ``place" actions across different primitive parts. This establishes a solid foundation for generalizable manipulation.

\textbf{ Language-Conditioned Skill Learning:} To learn the language-conditioned basic skill dataset, we design a transformer-based model $A$ based on the previous work~\cite{languagedeformable}.

\begin{figure}[htbp]
    \centering
    \includegraphics[width=\linewidth]{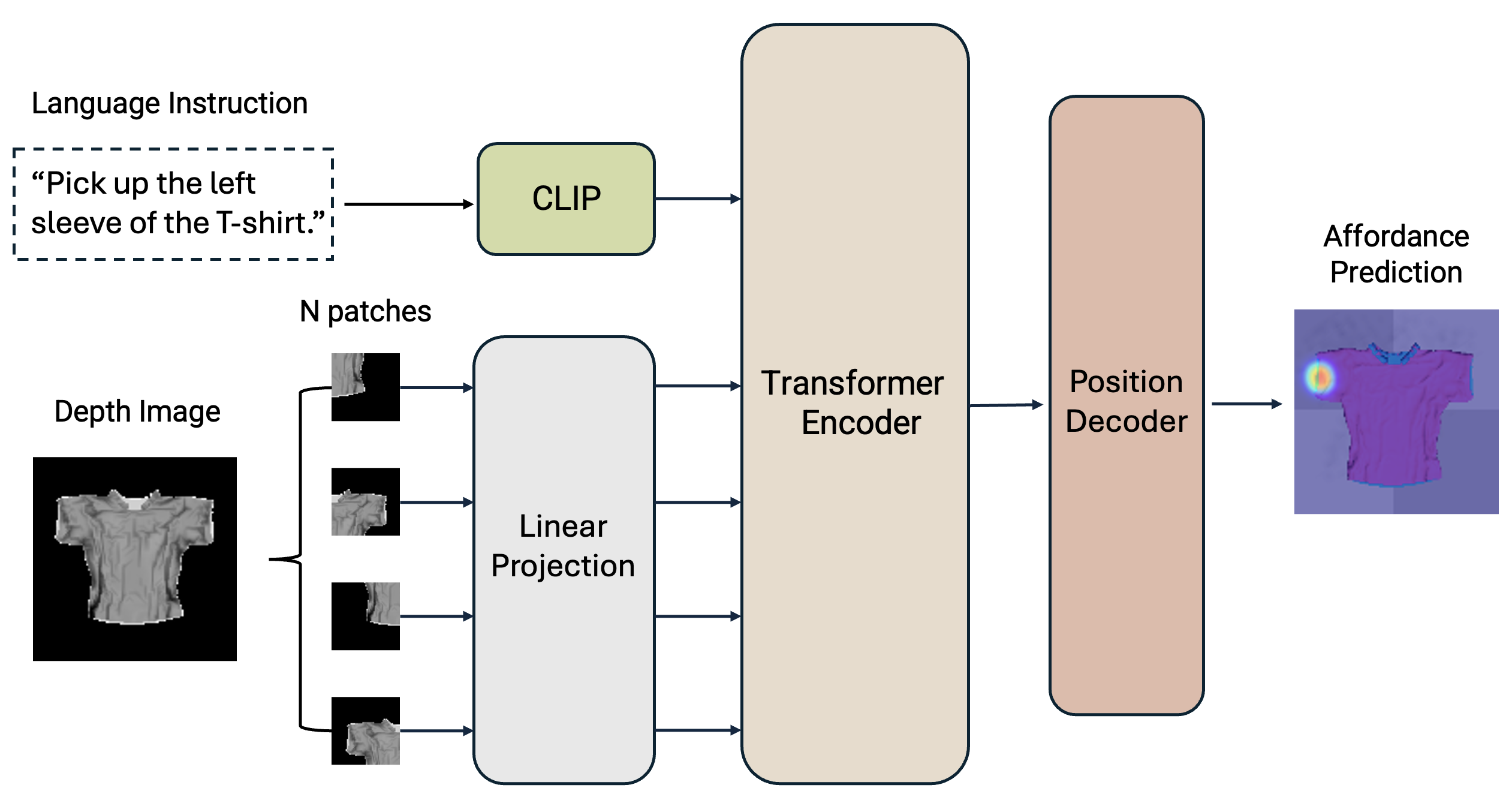}
    \caption{\textbf{Language-conditioned basic skills learning}. We train a Transformer-based model that takes language and depth images as input and outputs a heatmap of manipulation position.}
    \label{fig:learning}
    \vspace{-0.5cm}
\end{figure}

As shown in Fig.~\ref{fig:learning}, we first embed natural language instructions using the CLIP~\cite{radford2021learning} language encoder, enhancing the embeddings with position and learnable embeddings for context adaptability. For depth images, we employ a patch-based approach, dividing each image into $N$ patches, flattening them into 1D vectors, and applying linear projection. These vectors are augmented with position and learnable embeddings to preserve spatial information and improve adaptability. Next, modal-type embeddings are added, and the inputs are processed by a Vision Transformer (ViT)~\cite{dosovitskiy2020image} encoder. The encoder outputs are passed to a position decoder, consisting of alternating convolutional and upsampling layers, which generates a pixel-level affordance heatmap. This heatmap indicates the likelihood of each pixel being a ``pick'' or ``place'' point. The model selects the pixel with the highest likelihood as the target point for execution.\par

With model $A$, at time step $t$ in a multi-step manipulation task, the language instruction $l_t$ and corresponding depth image $o_t$ are mapped to a ``pick" or ``place" target point $p_t$ on the fabric. This target point is then converted into an arm trajectory through camera calibration and motion planning during skill execution.

\label{subsec:task_planning}
\subsection{Generalizable Cloth Manipulation through Composition of Learned Skills}
\textbf{High-level Planning through Large Language Model:} 
Given a new task described by an unseen language instruction $L_{new}$, we leverage an LLM to perform task planning to compose the existing basic skills learned from the long demonstrations. 
The instructions describe new or unanticipated combinations of actions, yet the LLM are able to handle them effectively. 
Using the same prompt structure as in the decomposition phase, the LLM is capable of breaking down these high-level instructions into individual, actionable language instructions $\text{LLM}(L_{new}) \to \{l_1, l_2, \dots, l_t\}$. 
For example, for a complex folding task, the LLM would break it down into multiple basic skills (e.g., \textit{``pick up the left sleeve of the T-shirt''} or \textit{``fold it to the left hem''}), each corresponding to a learned action step. This decomposition process enables the model to map abstract language commands to a sequence of tangible actions, each of which is linked to the learned single-step task $l_{ij}$. Consequently, the LLM is able to generate reasonable plans for unseen tasks. \par
\textbf{Skill Execution:} 
During the skill learning stage, we have already trained the model $A$ to ground the individual language instructions into real executable actions.
Once the long language instructions are decomposed into individual commands, these instructions are fed into the model. When input current observation $o_t$ and decomposed language $l_t$, the model predicts the corresponding ``pick'' or ``place'' point $p_t$. This prediction represents the target location in the image space where the robot arm needs to interact.
Given this 2D prediction interaction point in the pixel space $x_t=[px,py]^t$, the RGB-D image $I_t$, and the camera intrinsic $K$, we can directly compute the 3D interaction coordinates $X^c_t=[x,y,z]^t$ under the camera frame. Then, we transform the interaction coordinate to the base frame and combine the coordinate $X^b_t$ with primitive ``pick" or ``place" target point to compute the arm trajectory $T_t$. 
\label{subsec:action_generation}

\definecolor{lightblue}{rgb}{0.94, 0.97, 1}
\lstset{
    backgroundcolor=\color{lightblue},   % 背景颜色
    basicstyle=\footnotesize\ttfamily,   % 字体风格
    keywordstyle=\color{black},           % 关键词颜色
    commentstyle=\color{green},          % 注释颜色
    stringstyle=\color{brown},             % 字符串颜色
    numbers=left,                        % 行号
    numberstyle=\tiny\color{gray},       % 行号风格
    frame=single,                        % 边框
    breaklines=true,                     % 自动换行
    tabsize=4                            % Tab宽度
}

\begin{table*}[ht]
\centering
\caption{\textbf{Simulation experiment results}. The average success rates (\%) on seen and unseen multi-step cloth manipulation tasks of our model and two baselines.}
\begin{tabular}{lccccc}
\toprule[0.75pt]
\textbf{Task Type} & \textbf{Task} & \textbf{Cloth} & \textbf{Our Model} & \textbf{Language Deformable~\cite{languagedeformable}} & \textbf{Foldsformer~\cite{foldsformer}}  \\
\midrule
\multirow{5}{*}{Unseen} 
& Sleeves folding \& Half folding (vertical) & T-shirt & 98.0 & 0.0 & 44.0\\
& Four corners folding & Square & 100.0 & 0.0 & 26.0\\
& Half folding (vertical) & Skirt & 98.0 & 0.0 & 40.0\\
& Half folding (horizontal) & Rectangular & 98.0 & 0.0 & 62.0\\
& Half folding (vertical) & Trousers & 100.0 & 0.0 & 74.0\\

\midrule
\multirow{5}{*}{Seen} 
& Sleeves folding \& Half folding (horizontal) & T-shirt & 94.0 & 62.0 & 66.0\\
& One corner folding & Square & 100.0 & 100.0 & 100.0\\
& Half folding (horizontal) & Skirt & 92.0 & 100.0 & 68.0\\
& Half folding (vertical) & Rectangular & 100.0 & 100.0 & 90.0\\
& Half folding (horizontal) & Trousers & 96.0 & 100.0 & 84.0\\
\bottomrule[0.75pt]
\end{tabular}
\vspace{-0.5cm}
\label{tab:simulation comparison}
\end{table*}

\section{Experiments}

\label{sec:experiments}
We aim to answer the following research questions by presenting simulated and real-world experiments in this section: \par

\begin{table}[ht]
    \centering
    \caption{\textbf{Skill discovery performance of different foundation models}.}
    \label{tab:LLMs Test Result}  
    \begin{tabular}{lccc}
        \toprule
        \textbf{Model} & \textbf{Task 1} & \textbf{Task 2} & \textbf{Task 3} \\
        \midrule
        GPT-4o & 100\% & 100\% & 100\% \\
        Deepseek R1 & 100\% & 80\% & 60\% \\
        Qwen 2.5 Max & 60\% & 30\% & 0\% \\
        GPT-4o + Image & 100\% & 100\% & 70\% \\
        \bottomrule
    \end{tabular}
    \vspace{-0.1cm}
\end{table}

\begin{itemize}
\item[1)] How effectively does our method autonomously discover basic skills for cloth manipulation? (Sec.~\ref{subsec:skill discovery})
\item[2)] How well does our method perform on multi-step cloth
manipulation tasks compared with the existing baseline methods? (Sec.~\ref{subsec:simulation experiments})
\item[3)] Can our method generalize to unseen multi-step cloth
manipulation tasks? (Sec.~\ref{subsec:simulation experiments})
\item[4)] Can our method trained in simulation be successfully transferred to real-world cloth manipulation tasks? (Sec.~\ref{subsec:real experiments})
\end{itemize}

\subsection{Skill Discovery Experiments}
\label{subsec:skill discovery}
Autonomous basic skill discovery in cloth manipulation forms the foundation of our method. To this end, we implement the state-of-the-art large language models and vision language models, evaluating their performance on skill discovery. The selected models include ChatGPT-4o~\cite{achiam2023gpt}, Deepseek R1~\cite{guo2025deepseek}, and Qwen 2.5-Max~\cite{yang2024qwen2}.
 For GPT-4o, we assess its performance as both an LLM  and a VLM. When tested as a VLM, images of cloth are provided as supplementary information to aid in skill discovery. And for each model, we evaluate their performance on three multi-step tasks: Task 1, ``fold the trousers from left to right" (4 steps); Task 2, ``fold the right sleeve of the T-shirt inward, then fold the T-shirt from bottom to top" (6 steps); Task 3, ``Fold both sleeves inward, then fold the T-shirt top to bottom, then left to right" (12 steps). For each task, 10 trials are conducted to calculate the success rate of skill discovery, which is evaluated by human experts. The results are shown in Table~\ref{tab:LLMs Test Result}, indicating that GPT-4o outperforms the other models in terms of success rate. These findings demonstrate that GPT-4o exhibits strong reasoning abilities in skill discovery for cloth manipulation when used as an LLM. In subsequent experiments, we select GPT-4o as the LLM for skill discovery.

\begin{figure*}[htbp]
    \centering
    \includegraphics[width=\textwidth]{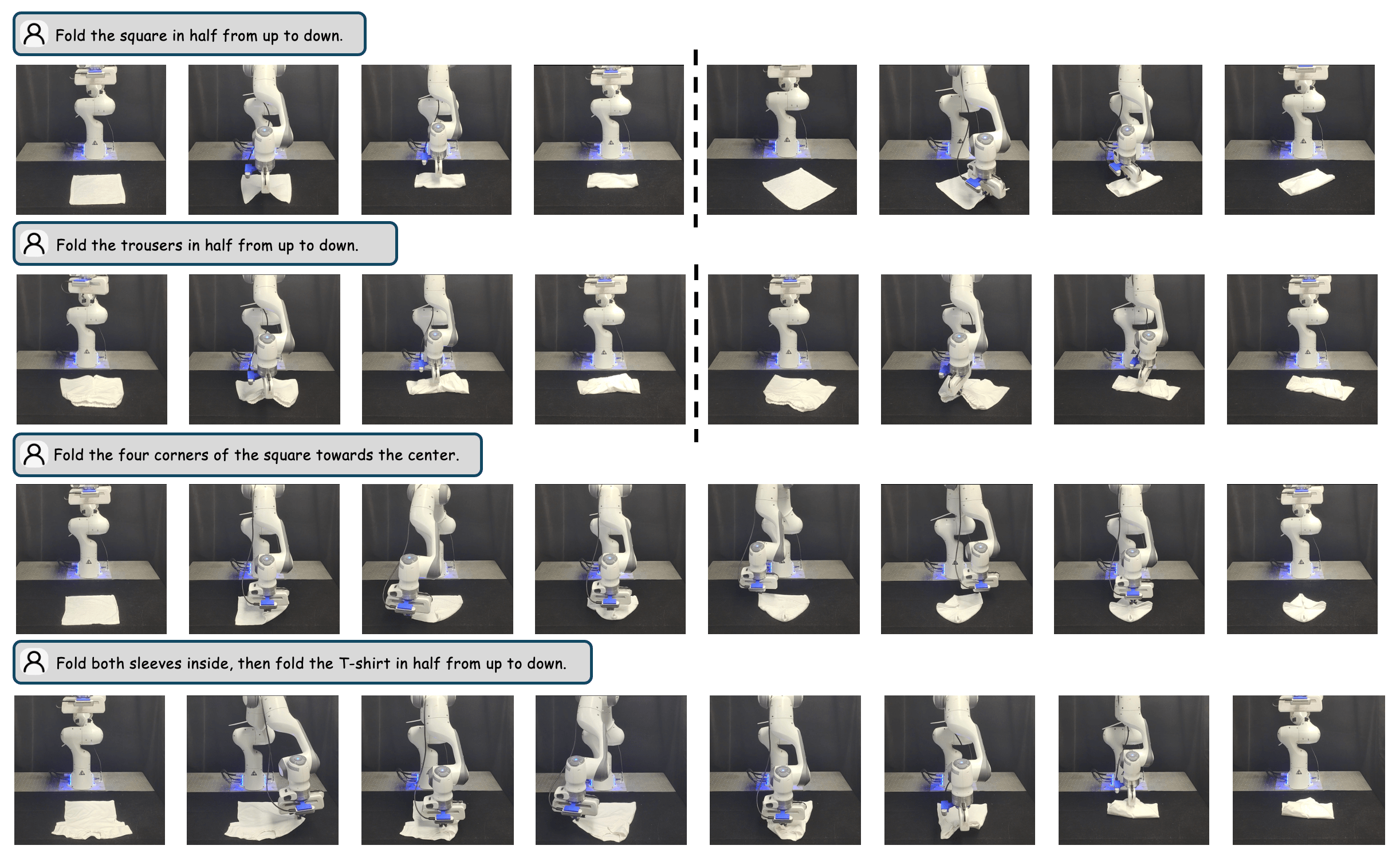}
    \caption{\textbf{Qualitative results of real experiments}. Our method performs well in multi-step manipulation tasks and can generalize to unseen tasks in the real world.}
    \label{fig:real_world_experiment}
    \vspace{-0.5cm}
\end{figure*}

\subsection{Simulation Experiments}
\label{subsec:simulation experiments}

To evaluate the performance of our method on multi-step cloth manipulation, we extend the SoftGym benchmark~\cite{lin2021softgym} to include five types of cloths: T-shirt, skirts, trousers, square, and rectangle cloth. The manipulation tasks are split into seen and unseen tasks. For all methods, seen tasks are provided during the model training stage, while unseen tasks are only presented during the testing stage. Unseen tasks require the model to generalize to the unseen folding requirement.\par

We compare our model with two multi-task learning baselines for multi-step cloth manipulation. Language-Deformable~\cite{languagedeformable} is an end-to-end algorithm for learning language-conditioned deformable object manipulation policies, which leverages a pretrained vision-language
model and a transformer architecture. Foldsformer~\cite{foldsformer} is a multi-step cloth manipulation learning algorithm,  which uses sequential demonstration images to specify different tasks and a space-time attention mechanism for long demonstration understanding and inference. We conduct experiments on 10 tasks, including 5 seen and 5 unseen tasks, with 50 trials for each task. A task is considered successful if the error between the policy's final result and the oracle's result is within 0.025m, which corresponds to the particle size of SoftGym. The results are shown in Table~\ref{tab:simulation comparison}.\par

The experimental results show that our model outperforms both baselines on both seen and unseen tasks. For seen tasks, the success rate of Language-Deformable is 100\% for all tasks except TshirtFold, while Foldsformer performs the worst. This indicates that natural language provides additional information beyond images in multi-task learning but fails to perform well when the task complexity is high. Our policy, on the other hand, learns basic skills, which are easier to learn compared to multi-step policies, resulting in a higher success rate on seen tasks.\par

For unseen tasks, the generalization of Language-Deformable is limited to simple spatial patterns, such as generalizing from ``left to right'' to ``right to left''. However, in our setting, a higher level of generalization is required, such as generalizing from ``vertical'' to ``horizontal'', where the action sequence is totally different. Foldsformer achieves a relatively higher success rate. However, for unseen tasks, it also requires complete sequential images from expert demonstrations as the task specification. These images provide sub-goals for multi-step learning, offering additional information that improves Foldsformer’s performance on unseen tasks. In contrast, our method does not require extra information from unseen tasks and achieves a much higher success rate.\par

\subsection{Real Experiments}
\label{subsec:real experiments}

\begin{table}[ht]
\centering
\Huge
\caption{\textbf{Quantitative results of real experiments}.}
\label{tab:real_word_experiment}
\renewcommand{\arraystretch}{1.5} % 设置行间距为原来的几倍
\resizebox{\columnwidth}{!}{%
\begin{tabular}{cccccc}
\toprule[2.8pt]
\multicolumn{1}{c}{\textbf{Task Type}} & \textbf{Task}  & \textbf{Cloth}  & \multicolumn{1}{c}{\textbf{MIoU}} & \multicolumn{1}{c}{\textbf{WR}} & \textbf{Success Rate}\\ \midrule
Seen & Half folding (vertical)     & Square               & 0.9153                          & 0.0134                         & 100\% \\ \midrule                                                                                                   
\multirow{3}{*}{Unseen}&Half folding (vertical)     & Trousers & 0.9103                           & 0.0325                         & 100\%\\                                                                                                     
&Corner folding               & Square   & 0.7970                          & 0.0301                        & 100\% \\                                                                                                            
&Sleeves folding\&Half folding (vertical) & T-shirt                   & 0.8978                           & 0.0189                        & 60\% \\ 
\bottomrule[2.8pt]
\end{tabular}
}
\vspace{-0.1cm}
\end{table}

\begin{figure}[htbp]
    \centering
    \includegraphics[width=\linewidth]{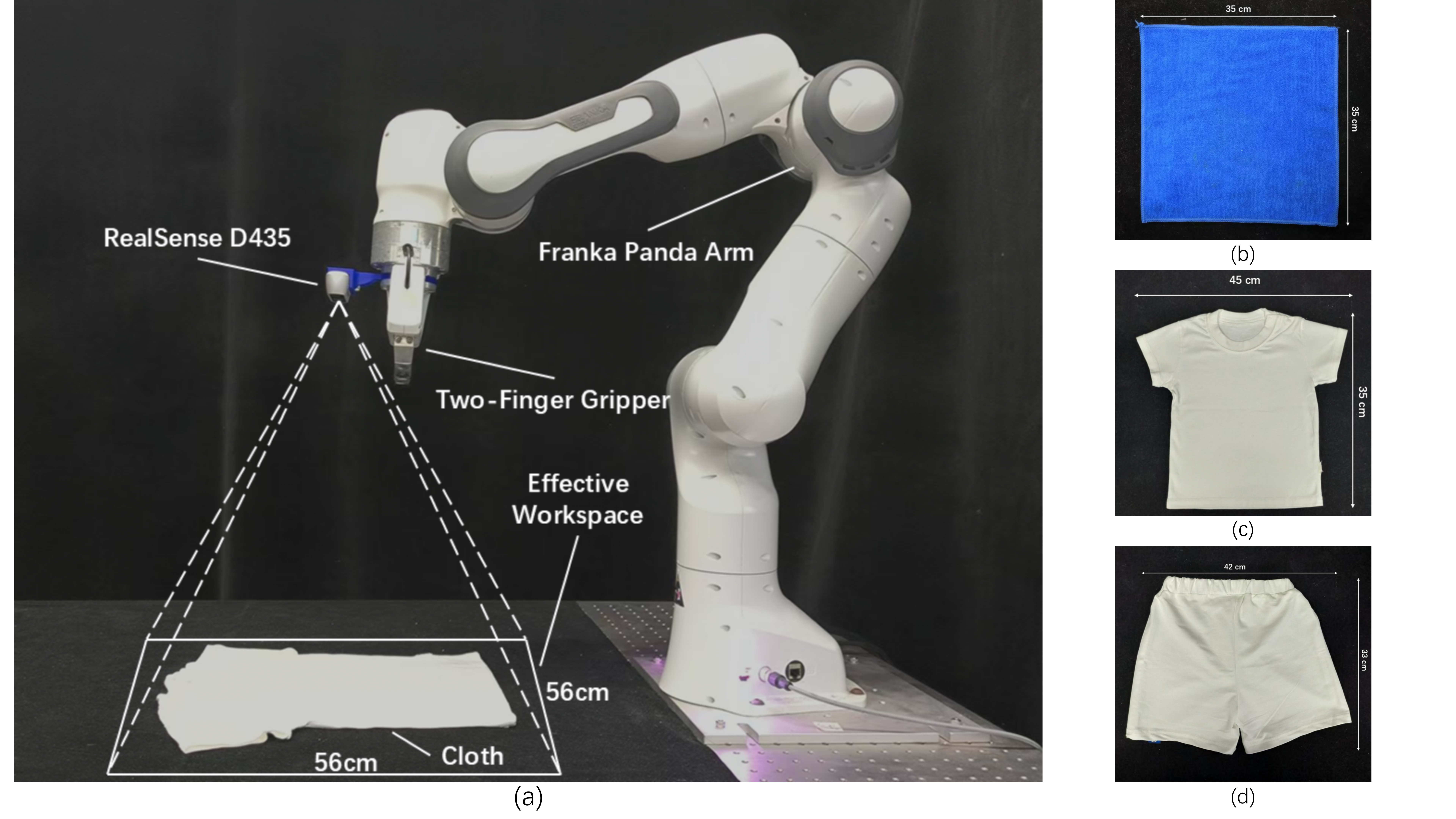}
    \caption{\textbf{Experimental setup}. (a) Robot system. 
(b) A 35cm × 35cm cloth. (c) A T-shirt (45cm × 35cm). (d) A pair of Trousers (42cm × 33cm).}
    \label{fig:setup}
    \vspace{-0.5cm}
\end{figure}

To evaluate the sim-to-real performance of our method, we conduct experiments in the real world using the Franka robot platform, with the setup shown in Fig.~\ref{fig:setup}. A Realsense RGB-D camera
is used to capture depth images of the cloth. We evaluate the performance of our method on multiple types of cloth manipulation tasks. A task is considered successful if the Mean Intersection over Union (MIoU) is greater than 0.8 and the Window Recall (WR) is less than 0.35. MIoU measures the overlap between the fabric states after human expert and robotic manipulation, while WR quantifies the presence of undesired edges and wrinkles.
Results are presented in Table~\ref{tab:real_word_experiment}. And Fig.~\ref{fig:real_world_experiment} illustrates some examples of our real-world experiments. The results indicate that our method can be effectively transferred to real-world scenarios, achieving high success rates on both seen and unseen tasks.\par

\section{Conclusion}
\label{sec:conclusion}
In this paper, we propose a novel pipeline for learning generalizable multi-step cloth manipulation skills. Leveraging the commonsense knowledge of the LLM, the proposed pipeline autonomously discovers and learns basic skills from long demonstrations and composes them to generalize to unseen tasks. Both simulation and real-world experiments demonstrate that the proposed pipeline is effective in multi-step cloth manipulation and generalizes well to unseen tasks. However, as the number of action steps increases, the observation may diverge significantly from the demonstrations used during model training. This distribution shift can result in failures in learned basic skills. In future work, we will explore more robust language-conditioned basic skill learning methods. One possible improvement is utilizing a pretrained large vision model as the feature encoder to enhance generalization to visual variations.\par
%% BIBLIOGRAPHY
\newpage
\bibliographystyle{IEEEtran}
\bibliography{bibliography.bib}
\end{document}